\documentclass{article}
\usepackage{spconf,amsmath,graphicx,cite}

\title{Personalized Query Rewriting in Conversational AI Agents}
\name{Alireza Roshan-Ghias, Clint Solomon Mathialagan, Pragaash Ponnusamy, Lambert Mathias, Chenlei Guo}
\address{Amazon Alexa, Seattle, WA, USA}

\begin{document}
\maketitle

\begin{abstract}
Spoken language understanding (SLU) systems in conversational AI agents often experience errors in the form of misrecognitions by automatic speech recognition (ASR) or semantic gaps in natural language understanding (NLU). These errors easily translate to user frustrations, particularly so in recurrent events e.g. regularly toggling an appliance, calling a frequent contact, etc. In this work, we propose a query rewriting approach by leveraging users' historically successful interactions as a form of \textit{memory}. We present a neural retrieval model and a pointer-generator network with hierarchical attention and show that they perform significantly better at the query rewriting task with the aforementioned user \textit{memories} than without. We also highlight how our approach with the proposed models leverages the structural and semantic diversity in ASR's output towards recovering users' intents.
\end{abstract}


\section{Introduction}
\label{sec:intro}
Advances in SLU technologies have fueled the growth of conversational AI agents such as Amazon Alexa, Google Home and Siri, where voice is the primary input modality. Dealing with ASR and NLU errors in these systems is crucial for providing a frustration-free customer experience. 

There has been a large body of work surrounding the detection and correction of ASR errors, which include either using contextual vocabulary during decoding \cite{Raju2018,Williams2018,Pundak2019}, augmenting speech recognition training datasets with implicit transcripts \cite{Shokouhi2016}, or rectifying historically prevalent errors via lattice rescoring \cite{Shivakumar2019}. All of these works focus on improving the ASR output. However, downstream components such as NLU, the dialog state manager, and skills can very well introduce errors, which are typically solved independently of each other. Moreover, some errors may be only recoverable in the context of the whole conversation. A holistic approach that bridges the learning across all components can decrease user friction, regardless of the source of error.

\begin{figure}[htb]
\begin{minipage}[b]{1.0\linewidth}
    \centering
    \includegraphics[width=1\textwidth]{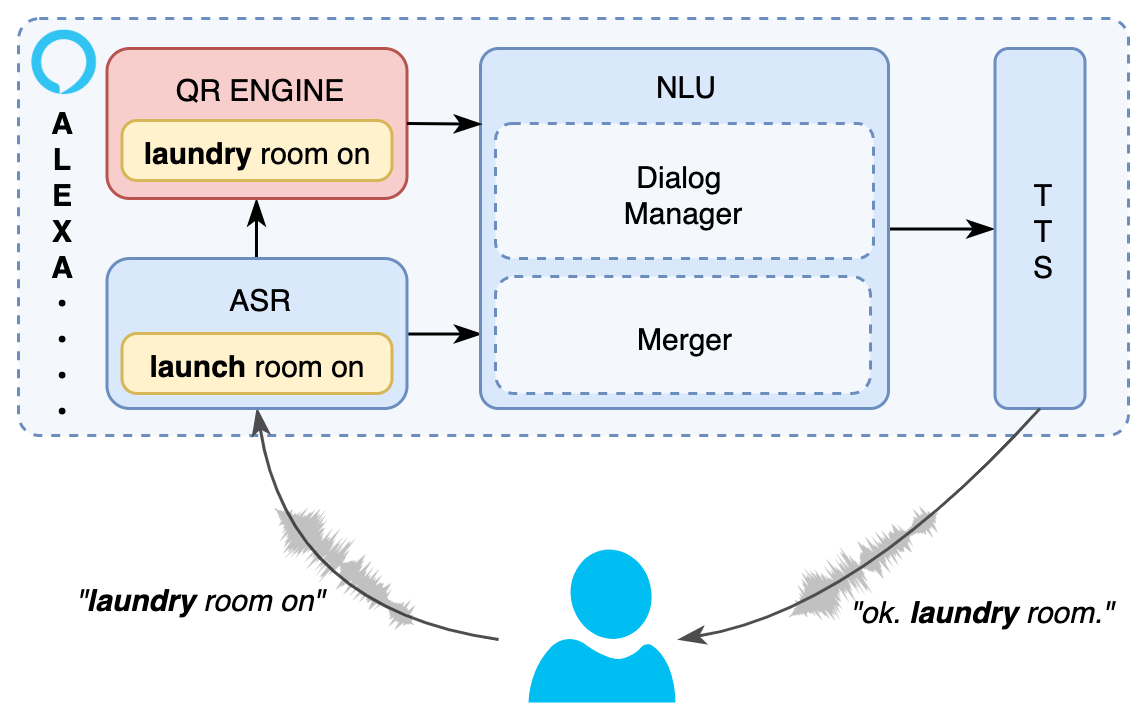}
    \caption{The query rewriting (QR) engine in context of the end-to-end SLU service.}
    \label{fig:architecture}
\end{minipage}
\end{figure}

Query rewriting has been successfully applied in search applications to increase recall, and/or correct misspelling by the users \cite{Baeza2005, dehghani2017learning}. Similarly, we propose resolving errors in SLU systems using query rewriting where the rewrite occurs on the utterance output of a high quality ASR system before being fed downstream into the NLU component. This reformulation strategy allows us to deal with errors across the entire pipeline in a single generalized framework. Similar to advances in dialog-based tasks that benefit from being grounded by dialogue contexts\cite{Yavuz2018},  knowledge bases \cite{Ghazvininejad2017,Mihail2017}, and users' persona \cite{Zhang2018,Mazare2018}, we hypothesize that the query rewriting task should also be grounded on users' interaction histories.

Figure \ref{fig:architecture} shows how the query rewriting engine fits into Alexa's architecture as a whole. To exemplify this, consider when the user says "\textit{laundry room on}", but ASR wrongly transcribes it as "\textit{launch room on}". By inspecting the user's history, we would know that the user never had a \textit{successful} utterance referencing "\textit{launch room}". As such, we propose rewriting it as "\textit{laundry room on}". Thereafter, both the original and the rewrite go through the NLU system in parallel before being arbitrated by a merger in rendering the final decision that indeed, the user actually meant to turn on the \textit{laundry room}.

\begin{figure}[htb]
\begin{minipage}[b]{1.0\linewidth}
    \centering
    \includegraphics[width=1\textwidth]{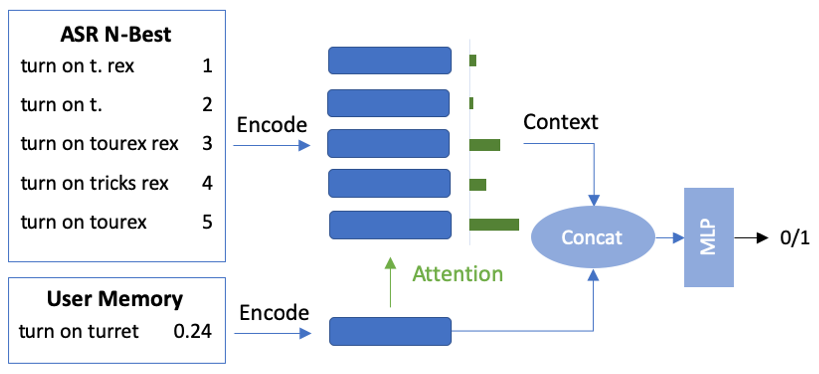}
    \caption{Retrieval model with user memory. Blue and green rectangles denote encoded vectors and attention, accordingly.}
    \label{fig:retrieval}
\end{minipage}
\end{figure}
In this paper, we propose a personalized query rewriting approach by leveraging user \textit{memories} to reduce errors in a large-scale SLU system. Our main contributions are: 

\begin{enumerate}
    \item We present a novel approach for query rewriting that leverages user-specific memories aggregated based on their \textit{successful} historical interactions.
    \item We compare and contrast two neural architectures - a retrieval based approach as well as a generative approach on our task.
    \item We demonstrate the impact of leveraging ASR uncertainty in query rewriting.
\end{enumerate}



\section{Data}
\label{sec:data}
In this work, we sample a subset of anonymized user interactions with Alexa (complying with user instructions on data retention) for our experiments, focusing specifically on smart home, where there is a need to understand the set of personalized appliances each user has.

Our data consists of two parts: User memories and rephrase pairs. For user memories, we mine and aggregate queries over 4 weeks of the \textit{successful}\footnote{Defined as an interaction that Alexa executed for the user and the user did not initiate a follow-up intent in that domain for a specified period of time. For example, the air conditioner was set to a setting, or a device was turned off, and the user did not change them for a given period of time} dialogues. For rephrase pairs, we mine pairs of smart home utterances across Alexa users in which the first utterance\footnote{We use the n-best output from the ASR system, and so refer to this as {\it ASR n-best}} is defective, i.e. resulted in an Alexa error, and the second consecutive successful utterance\footnote{We refer to this as a {\it rephrase}}. In order to minimize overlap, and make the comparisons fair, the rephrase data is collected for the week after the user memory is aggregated. Since some users might change their mind in the second turn, or they might issue a new command, the second utterance may not semantically match a memory utterance. We consider these cases as "non-rewritable". A semantic match is defined on the intent and slot output of the NLU system. This allows us to declare, for example, "turn on the light" and "can you turn on light" as a semantic match. We opted to not remove the non-rewritable pairs from our training data, since we want to \textit{learn} to not rewrite these cases.  Note that we do not need any "annotations" to establish our training and testing dataset.

We use top 5 ASR n-best for the first utterance in the rephrase pair to incorporate ASR uncertainty. We finally joined these two datasets ($\sim$1M data points), and split them by randomly assigning 80\% of users to training, and the rest to the testing set.

\section{Models}
\subsection{Baseline models}
\subsubsection{Pointer Network without User Memory}
\begin{figure*}[t]
    \centering
    \includegraphics[width=1.0\textwidth]{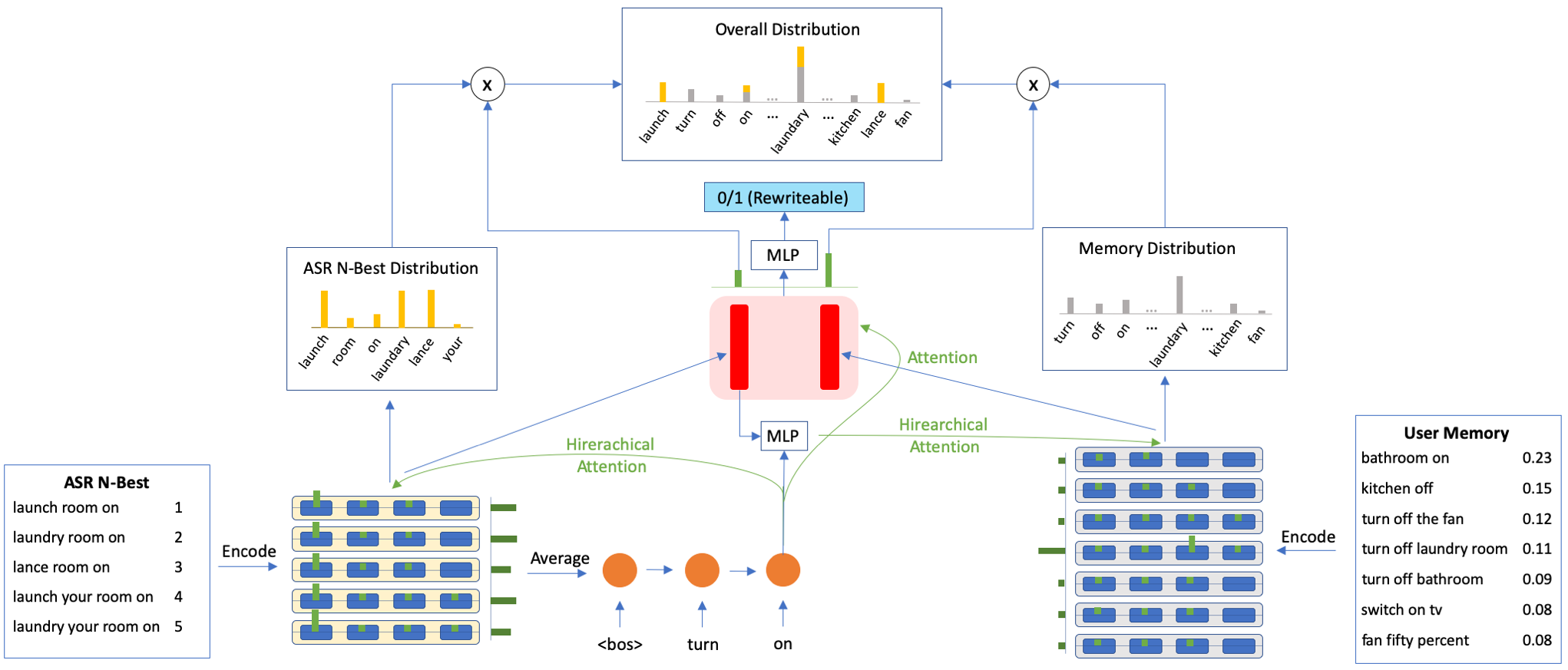}
    \caption{Memory-based pointer network model. Blue, red and green rectangles denote word-level encoder outputs, context vectors, and attention weights, respectively. Frequency of each user memory is concatenated with the word embedding before passing through the encoder. Note that the \textit{rewritable flag} is either 1 or 0 for the entirety of the sequence.}
    \label{fig:seq2seq}
\end{figure*}

In order to evaluate the impact of having user memories on the performance of query rewriting, we first establish a baseline without memory. This approach is akin to learning to "re-rank" the ASR n-best using the rephrase dataset. The model is similar to the pointer network model (Figure \ref{fig:seq2seq}) without the memory related components. For more details, we refer the reader to Section~\ref{ssec:pgn}.

\subsubsection{Retrieval Baseline with User Memory}
We use a point-wise one-way attention for the retrieval model \cite{Guo2019}. Figure \ref{fig:retrieval} shows the architecture of the model. We encode each ASR n-best and memory using bi-LSTM, then use the encoded memory to attend on the ASR n-best, and calculate ASR n-best context accordingly. We then concatenate the encoded memory and the ASR n-best context, and pass it through two dense layers followed by a sigmoid. The label is 1, if the rewrite matches the rephrase. The model is trained using cross-entropy loss. Each memory (represented as an utterance) is encoded independently, and the highest scoring memory is retrieved as the final query rewrite.

To encode the tokens at the word level, we sum up the associated subword embeddings. Subwords are determined using byte-pair encoding \cite{Sennrich2016}. The subword embeddings are learnt as part of the training process.

\subsection{Pointer Network with User Memory}
\label{ssec:pgn}
We extend the pointer networks architecture \cite{See2017, Yavuz2018} by implementing 1) hierarchical attention on both ASR n-best and user memory, and 2) multi-task learning to both generate an utterance and predict whether it is "rewritable" or not (Figure \ref{fig:seq2seq}). We first encode all utterances using a bidirectional LSTM~\cite{Hochreiter:1997:LSM:1246443.1246450}. Similar to the retrieval model, word embedding is the sum of each word's subword embeddings. At each decoding step, the final word probability distribution is calculated as a mix of vocabulary probability distributions from user memory and ASR n-best, each corresponding to the sum of hierarchical attention per word \cite{Yavuz2018}. After calculating the final vocabulary distribution, we pick the word with maximum probability and pass it back to the decoder as the input for the next decoding step. In order to predict whether to rewrite or not, we use the ASR n-best and user memory contexts to predict whether we should rewrite or not at every timestep. To train the model, we combine the cross-entropy loss of pointer network part and the binary loss for the rewritable prediction part. We determine the mixing ratio using hyperparameter optimization. We also mask the loss for the non-rewritable utterances for the pointer network part.

\section{Results}
\label{sec:results}

\subsection{Evaluation Criteria}
In order to evaluate the results, we compare the predicted rewrite and the actual rephrase for a given ASR n-best. In the retrieval model, the highest point probability for each user memory is considered as the "rewrite probability". In the pointer network model, we have two arrays of probabilities corresponding to each timestep's word probability and rewritability probability. We calculate the geometric mean of each array, and again, take the geometric mean of the those means to represent the "rewrite probability". This way we are combining the confidence in the rewrite and rewritability prediction, as estimated by the model.

We define true positive (TP) as when the rewrite matches the rephrase. False positive (FP) is when rewrite is not equal to the rephrase in both rewritable and non-rewritable cases. False negative (FN) is defined only for rewritable utterances that the model doesn't rewrite (due to low confidence). True negative (TN) is defined when the utterance is non-rewritable and we don't rewrite it.We also measure the performance of each approach in cases where the \textit{intent} of the rephrase and rewrite are different.

\subsection{Experiments}
\subsubsection{Training Setup}
All models are implemented in PyTorch~\cite{paszke2017automatic}. Sub-word embedding dimensions in all models were set to 100, and were randomly initialized and updated during the training. We set the size of encoder and decoder LSTM hidden layers to 100. Since we are concatenating frequency of occurrence with word embeddings, we have two different encoders for n-best and user memory. We used Adam optimizer with batch sizes of 512 and 256 for the retrieval and pointer network models, respectively, with default learning rates. We use greedy decoding for the pointer network model. All models are trained on a single GPU p3.2x large AWS instance.



\subsubsection{Results}
For the retrieval model, we tried three different encoders: 1) Average of all word embeddings, 2) passing each utterance through bi-directional LSTM, and taking the average of encoder outputs, and 3) self-attention on the bi-LSTM encoder outputs \cite{Lin2017}. We also experimented with both multiplicative and additive attentions \cite{Bahdanau2015}. We found additive attention to be marginally better.

\begin{table*}
        
        \centering
        \begin{tabular}{ l c c c c }
            
            \hline
            \bf{Model} & \bf{N-Best} & \bf{Memory} & $\textbf{AUC}_{PR}$ & \bf{Recall} \\
            \hline \hline
            Pointer Network & 5 & No & 0.225 & N/A \\
            \hline
            Retrieval & 1 & Yes & 0.813 & 19.7 \\
            Pointer Network & 1 & Yes & 0.821 & 28.4 \\
            \hline
            Retrieval & 5 & Yes & 0.848 & 36.6 \\   
            \bf{Pointer Network} & 5 & Yes & \bf{0.856} & \bf{42.0} \\
            \hline

        \end{tabular}
   
    \caption{Comparing the three models based on area under precision-recall curve (PR-AUC) and recall at precision 90\%.}
   
    \label{tab:results}
   

\end{table*}

For the pointer network model, we experimented with different attentions (additive vs. multiplicative), and slight changes to the architecture. Important factors were frequency of user memory, and how context was defined. We found that if we only use the decoder output to attend to memories, the results are significantly weaker. The best results were obtained when decoder output was combined with ASR n-best context to attend on user memories. Other alternatives such as attending to the memory first, and using the context to attend to ASR n-best proved less effective. This can be due to the fact that number of memory utterances are usually higher than ASR n-best, so it is easier to get the correct context from ASR n-best.

As expected, the pointer network model without memory performs poorly (Table \ref{tab:results}). The model is not even able to achieve precision of 90\%, since it produces many false positives. This was expected, since without memory, it is difficult to know if the user meant "bathroom", or "bedroom", for example. Only when we look at the user memory and we see that the user only has turned on and off bedroom, then we can rewrite to "turn on bedroom". Both retrieval and pointer network models perform significantly better with memory, with the pointer network model having an edge, especially at having a higher recall when we fix the precision to 90\%. We also observed that the pointer network model is 5 times less likely to get the intent wrong. This shows that the pointer network model is learning to effectively construct the right utterance. For example, it learns to deal with patterns like "turn on no turn off the fan" by taking the latter intent from the utterance. Another example of how the pointer network model excels is when the user memory only contains "turn on bedroom" intent, for example, and what the user want is to "turn off bedroom". The retrieval model can mistakenly retrieve the opposite intent, but the pointer network model is able to "compose" the right intent by copying "turn off" from the ASR n-best, and the device name from the memory.

It is also clear that adding ASR uncertainty (in the form of n-best) helps with the performance for both retrieval and pointer network models. This is not a surprise, since it provides more signals for the model to learn from. There are even cases (like the example in Figure \ref{fig:seq2seq}) that what the user meant is already included in the n-best ($n > 1$).

Also, note that since our test data is not annotated, there is a minority of cases where the rephrase (i.e. second turn) presents a different intention compared to the ASR n-best (i.e. first turn). For example, we might have "turn on tv" as the first turn, and per definition, it didn't succeed, and in the second turn, the user changes his/her mind and says "turn on the lamp". This results in not knowing the upper-bound of precision-recall for the model. However, since we are comparing the methods on the same metric and on the same test set, this noise is cancelled out.

\section{Conclusion}

In this paper, we proposed a personalized query rewriting approach using user-specific \textit{memories} to reduce frictions associated with Alexa interactions. This is achieved by conditioning the rewrite on both the ASR n-best and the user's \textit{memories}. We have shown that the pointer-generator network architecture was able to outperform the retrieval-based baseline model. We note that this is particularly interesting especially since the generative model takes on a more complex approach to query rewriting than its retrieval counterpart. Here, the generative model would need to learn an implicit language model to reconstitute the utterance exactly as the rephrase, while the retrieval model focuses squarely on structural and semantic matching against the user \textit{memory} entries.  

By taking on a more complex task, the pointer network also learns to get the intent right, far better than the retrieval model. The pointer network also has the capability to rewrite only a part of the input query. Consider an ASR output of "set the \textit{landry} room to \textit{white}". The pointer model can learn to rewrite this to "set the \textit{laundry} room to \textit{white}" even if the user memory contains "set the \textit{laundry} room to \textit{yellow}". In contrast, a retrieval model relies on having the exact rephrase in the user memory. While the pointer network is end-to-end and works with all the available memory simultaneously, the retrieval model only works with one memory at a time, and has a separate ranking logic. However, the pointer network has a higher latency compared to the retrieval model.



The proposed framework is easily extensible to other sources of knowledge and memory such as users' current conversation history. This flexibility allows for more complex conversation-level query rewriting. As a future work, we plan on incorporating dialogue context similar to the architectures proposed in~\cite{rastogi2019scaling}. Finally, our proposed method also takes full advantage of ASR n-best without invoking any additional overhead on downstream pipelines. In the future, we will explore using the ASR lattice output directly, as it provides a richer and more diverse hypothesis space.




\vfill\pagebreak

\bibliographystyle{IEEEbib}
\bibliography{refs.bib}

\begin{thebibliography}{10}

\bibitem{Raju2018}
Anirudh Raju, Behnam Hedayatnia, Linda Liu, Ankur Gandhe, Chandra Khatri,
  Angeliki Metallinou, Anu Venkatesh, and Ariya Rastrow,
\newblock ``{Contextual Language Model Adaptation for Conversational Agents},''
\newblock in {\em Interspeech}, 2018.

\bibitem{Williams2018}
Ian Williams, Anjuli Kannan, Petar Aleksic, David Rybach, and Tara~N. Sainath,
\newblock ``{Contextual speech recognition in end-to-end neural network systems
  using beam search},''
\newblock in {\em Interspeech}, 2018, pp. 2227--2231.

\bibitem{Pundak2019}
Golan Pundak, Tara~N. Sainath, Rohit Prabhavalkar, Anjuli Kannan, and Ding
  Zhao,
\newblock ``{Deep Context: End-to-end Contextual Speech Recognition},''
\newblock in {\em 2018 IEEE Spoken Language Technology Workshop, SLT 2018 -
  Proceedings}, 2019, pp. 418--425.

\bibitem{Shokouhi2016}
Milad Shokouhi, Umut Ozertem, and Nick Craswell,
\newblock ``{Did you say U2 or Youtube? Inferring Implicit Transcripts from
  Voice Search Logs},''
\newblock in {\em World Wide Web Conference}, 2016.

\bibitem{Shivakumar2019}
Prashanth~Gurunath Shivakumar, Haoqi Li, Kevin Knight, and Panayiotis Georgiou,
\newblock ``{Learning from past mistakes: Improving automatic speech
  recognition output via noisy-clean phrase context modeling},''
\newblock {\em APSIPA Transactions on Signal and Information Processing}, vol.
  8, 2019.

\bibitem{Baeza2005}
Ricardo Baeza-Yates, Carlos Hurtado, and Marcelo Mendoza,
\newblock ``{Query Recommendation Using Query Logs in Search Engines},''
\newblock in {\em International Conference on Extending Database Technology},
  2005, pp. 588--596.

\bibitem{dehghani2017learning}
Mostafa Dehghani, Sascha Rothe, Enrique Alfonseca, and Pascal Fleury,
\newblock ``Learning to attend, copy, and generate for session-based query
  suggestion,''
\newblock in {\em Proceedings of the 2017 ACM on Conference on Information and
  Knowledge Management}. ACM, 2017, pp. 1747--1756.

\bibitem{Yavuz2018}
Semih Yavuz, Abhinav Rastogi, Guan-lin Chao, and Dilek Hakkani-T{\"{u}}r,
\newblock ``{DEEPCOPY: Grounded Response Generation with Hierarchical Pointer
  Networks},''
\newblock in {\em NIPS Conversational AI workshop}, 2018.

\bibitem{Ghazvininejad2017}
Marjan Ghazvininejad, Chris Brockett, Ming-Wei Chang, Bill Dolan, Jianfeng Gao,
  Wen-tau Yih, and Michel Galley,
\newblock ``{A Knowledge-Grounded Neural Conversation Model},''
\newblock in {\em AAAI}, 2017, pp. 5110--5117.

\bibitem{Mihail2017}
Eric Mihail, Laksmi Krishnan, Francois Charette, and Christopher~D. Manning,
\newblock ``{Key-Value Retrieval Networks for Task-Oriented Dialogue},''
\newblock in {\em Annual SIGdial Meeting on Discourse and Dialogue}, 2017.

\bibitem{Zhang2018}
Saizheng Zhang, Emily Dinan, Jack Urbanek, Arthur Szlam, Douwe Kiela, and Jason
  Weston,
\newblock ``{Personalizing Dialogue Agents: I have a dog, do you have pets
  too?},''
\newblock in {\em ACL}, 2018, pp. 2204--2213.

\bibitem{Mazare2018}
Pierre-Emmanuel Mazar{\'{e}}, Samuel Humeau, Martin Raison, and Antoine Bordes,
\newblock ``{Training Millions of Personalized Dialogue Agents},''
\newblock in {\em EMNLP}, 2018, pp. 2775--2779.

\bibitem{Guo2019}
Jiafeng Guo, Yixing Fan, Liang Pang, Liu Yang, Qingyao Ai, Hamed Zamani, Chen
  Wu, W~Bruce Croft, and Xueqi Cheng,
\newblock ``{A Deep Look into neural ranking models for information
  retrieval},''
\newblock {\em Information Processing and Management}, 2019.

\bibitem{Sennrich2016}
Rico Sennrich, Barry Haddow, and Alexandra Birch,
\newblock ``{Neural Machine Translation of Rare Words with Subword Units},''
\newblock in {\em ACL}, 2016, pp. 1715--1725.

\bibitem{See2017}
Abigail See, Peter~J. Liu, and Christopher~D. Manning,
\newblock ``{Get To The Point: Summarization with Pointer-Generator
  Networks},''
\newblock in {\em ACL}, 2017.

\bibitem{Hochreiter:1997:LSM:1246443.1246450}
Sepp Hochreiter and J\"{u}rgen Schmidhuber,
\newblock ``Long short-term memory,''
\newblock {\em Neural Comput.}, vol. 9, no. 8, pp. 1735--1780, Nov. 1997.

\bibitem{paszke2017automatic}
Adam Paszke, Sam Gross, Soumith Chintala, Gregory Chanan, Edward Yang, Zachary
  DeVito, Zeming Lin, Alban Desmaison, Luca Antiga, and Adam Lerer,
\newblock ``Automatic differentiation in {PyTorch},''
\newblock in {\em NIPS Autodiff Workshop}, 2017.

\bibitem{Lin2017}
Zhouhan Lin, Minwei Feng, Cicero~Nogueira dos Santos, Mo~Yu, Bing Xiang, Bowen
  Zhou, and Yoshua Bengio,
\newblock ``{A Structured Self-attentive Sentence Embedding},''
\newblock in {\em ICLR}, 2017, pp. 1--15.

\bibitem{Bahdanau2015}
Dzmitry Bahdanau, KyungHyun Cho, and Yoshua Bengio,
\newblock ``{Neural Machine Translation by Jointly learning to Align and
  Translate},''
\newblock in {\em ICLR}, 2015.

\bibitem{rastogi2019scaling}
Pushpendre Rastogi, Arpit Gupta, Tongfei Chen, and Lambert Mathias,
\newblock ``Scaling multi-domain dialogue state tracking via query
  reformulation,''
\newblock in {\em Proceedings of the 2019 Conference of the North American
  Chapter of the Association for Computational Linguistics}. 2019, Association
  for Computational Linguistics.

\end{thebibliography}

\end{document}